\documentclass[journal,comsoc]{IEEEtran}
\usepackage{amssymb}
\usepackage{amsthm,amssymb}
\usepackage{eucal}
 \usepackage{multicol}

 \usepackage{graphicx}
\usepackage{cite}
\usepackage{amsmath,amssymb,amsfonts}
\usepackage{amsmath}
\usepackage{algorithmic}
\usepackage{mathtools}  
\usepackage{diffcoeff}  
\usepackage{graphicx}
\usepackage{textcomp}
\usepackage{xcolor}
\usepackage{lettrine}
 \usepackage{multirow}
 \usepackage{amsmath}
 \usepackage{graphicx}
 \usepackage{textgreek}
\usepackage{amssymb}
\usepackage{caption}
\usepackage{subcaption}
\usepackage{authblk}
\usepackage{tikz,lipsum}
\newcommand{\RN}[1]{%
  \textup{\uppercase\expandafter{\romannumeral#1}}%
}
\usepackage[ruled,longend]{algorithm2e}
\def\BibTeX{{\rm B\kern-.05em{\sc i\kern-.025em b}\kern-.08em
    T\kern-.1667em\lower.7ex\hbox{E}\kern-.125emX}}
\def\BibTeX{{\rm B\kern-.05em{\sc i\kern-.025em b}\kern-.08em
    T\kern-.1667em\lower.7ex\hbox{E}\kern-.125emX}}
\newtheorem{proposition}{\textbf{Proposition}}

\begin{document}

\title{ Recurrent Neural Network-based Anti-jamming Framework for Defense Against Multiple Jamming Policies }
\IEEEoverridecommandlockouts

  \author{\IEEEauthorblockN{\normalsize Ali Pourranjbar,    Georges Kaddoum,  \textit{Senior Member, IEEE} and Walid Saad, \textit{Fellow, IEEE}
}
\thanks{
A. Pourranjbar and G. Kaddoum are with the LaCIME Lab, Department of Electrical Engineering, École de technologie supérieure, Montreal,
QC H3C 0J9, Canada (e-mail: ali.pourranjbar.1@ens.etsmtl.ca; Georges.Kaddoum@etsmtl.ca).\\Walid Saad is with Wireless@VT, Bradley Department of Electrical and Computer Engineering, Virginia Tech, Blacksburg,
VA 24061 USA (e-mail:  walids@vt.edu). This article was presented in part at the
Asilomar conference on signals, systems, and computers in California, USA \cite{pourranjbar2021jamming}.

}
 }
\maketitle

\begin{abstract} 
Conventional anti-jamming methods mainly focus on preventing single jammer attacks with an invariant jamming policy or jamming attacks from multiple jammers with similar jamming policies. These anti-jamming methods are ineffective against single jammer following several different jamming policies or  multiple jammers with distinct policies. Therefore, this paper proposes an anti-jamming method that can adapt its policy to  current jamming attack. Moreover, for the multiple jammers scenario, an anti-jamming method that estimates the future
occupied channels using the jammers' occupied channels in previous time slots is proposed. In both single and multiple jammers scenarios, the interaction between the users and jammers is modeled using recurrent neural networks (RNN)s. The performance of the proposed anti-jamming methods is evaluated by calculating the users' successful transmission rate (STR) and ergodic rate (ER), and compared to a baseline based on Q-learning (DQL). Simulation results show that for the single jammer scenario,  all the considered jamming policies are perfectly detected  and high STR and ER  are maintained. Moreover, when $70\%$ of the spectrum is under jamming attacks from multiple jammers, the proposed method achieves a  STR  and ER greater than $75\%$ and $80\%$, respectively. These  values rise  to $90\%$ when $30\%$ of the spectrum is under jamming attacks.
In addition, the proposed anti-jamming methods significantly  outperform the DQL method   for all the considered cases and jamming scenarios.
\end{abstract}

 \vspace{-0.3cm}
\section{Introduction}
Wireless communication networks are susceptible to jamming attacks due to their shared and open nature. Jamming attacks cause performance degradation or denial of service by disrupting communication links in wireless networks. Therefore, it is necessary to adopt  anti-jamming methods to mitigate such attacks.
Various anti-jamming techniques have been proposed in the literature; however, the majority of available techniques is focused on preventing a single type of jamming attack. For instance, the authors in  \cite{ xuan2011trigger,nan2020mitigation,tcomkhodam,d2014defeating} mainly  focus on mitigating  reactive jammer attacks while \cite{slimeni2018modified} and \cite{yao2019collaborative} perform  anti-jamming against sweeping jammers. Some machine learning-based anti-jamming techniques  that can work against several jammer types, such as \cite{elleuch2021novel}, assume that during interactions between legitimate users and jammers, the policies of the jammers remain unchanged. Thus, if the policy of a jammer changes, legitimate nodes must be retrained, resulting in a performance loss. Therefore, there is a pressing need for  an anti-jamming method capable of mitigating multi-type jammer attacks.

When confronted with a single jammer that can adopt different jamming policies, it is  necessary to first recognize the jammer's policy, and then select an appropriate countermeasure. 
Jamming recognition techniques are mainly studied in the context of radar \cite{qu2020jrnet,shao2020convolutional ,wu2017jamming, wang2019recognition}, with a focus on detecting the jammer's type utilizing the jamming signals.
 In \cite{qu2020jrnet}, the power spectrum of the jamming signal  is used to determine the type the jammer. In \cite{shao2020convolutional}, the authors utilize the fast Fourier transform (FFT) of the jamming signal to recognize the jamming policy while in \cite{wu2017jamming} and \cite{wang2019recognition} the  time domain signal is used. The authors in \cite{cai2019jamming} employ convolutional neural networks (CNNs) to recognize the jamming type using a waterfall plot of the spectrum.

From a practical perspective, implementing the  recognition techniques from \cite{shao2020convolutional ,wu2017jamming, wang2019recognition, qu2020jrnet} requires accurate samples from the jamming signal. Moreover, the recognition technique in \cite{cai2019jamming} is limited to simple jamming policies and requires a large data set  for training. Thus, it is necessary to develop an anti-jamming technique that can work with a small data set for training and can be extended to different types of jamming. 
 
In addition, the presence of multiple jammers  with different jamming policies gives rise to additional
challenges in wireless networks. However, most   previous works, such as  \cite{xuan2011trigger,nan2020mitigation,tcomkhodam,d2014defeating,slimeni2018modified,yao2019collaborative}, consider the   single jammer case.  Some works,  such as  \cite{garnaev2020multi,garnaev2021multi,van2020defeating,  lotfi2021protecting} consider   multiple jammers with similar  jamming policies. Specifically, the authors in \cite{garnaev2020multi} and \cite{garnaev2021multi}  formulate the interaction between a user and several jammers as a non-cooperative game where the utility functions of the jammers are the same.  In \cite{garnaev2020multi} the latency is considered as the utility function of the  jammers while in \cite{garnaev2021multi} the SINR is set as the  jammers' utility function.  The authors in \cite{van2020defeating} propose to harvest  the reactive jammers' energy and use backscatter to relay the user's data. Similarly, the work in \cite{lotfi2021protecting} considers  a multi-function wireless system and employs a  backscatter as a redundant communication path when the main path is under threat from reactive jammers. Moreover, the works in \cite{le2018multiple,bhamidipati2018simultaneous,juhlin2021localization}  focus on the localization of the multiple jammers and do not propose any anti-jamming methods. In the context of Unmanned aerial vehicles (UAVs), the works in  \cite{li2021uav,wu2020energy,wu2021uav} attempt to  determine the trajectory for  UAVs that are under attack by multiple jammers to avoid jamming attacks. In \cite{paperkhodam} and \cite{lunden2015distributed}, the authors propose  a channel allocation method based on  spectrum sensing. In this context, the users share their local spectrum sensing with each other to select the channels that are less probable to be jammed.

 Although the works  in  \cite{van2020defeating,garnaev2020multi,garnaev2021multi,lotfi2021protecting,le2018multiple,bhamidipati2018simultaneous,juhlin2021localization,li2021uav,wu2020energy,wu2021uav,lunden2015distributed,paperkhodam} study scenarios with  multiple jammers, they
suffer from several drawbacks, which make them impractical against multiple jammers with
different jamming policies. The assumption of similar policies at the jammers  is a major drawback in all the above  mentioned works. Moreover, some of these works such as \cite{garnaev2020multi} and \cite{garnaev2021multi}, assume the availability of the channel gains between the user and jammers, which is not practical. The considered schemes in \cite{van2020defeating} and \cite{lotfi2021protecting} are restricted to specific system models because a backscatter is employed to relay the users' data in case of jamming attacks. The proposed solutions in  \cite{li2021uav,wu2020energy,wu2021uav} can only be applied  to UAVs, and therefore is unsuited for a wider range of applications. The authors in \cite{paperkhodam} and \cite{lunden2015distributed} propose to monitor  the jammers' occupied channels in previous time slots to allocate free channels to  the users in future time slots. This approach is practical against multiple jammers with multiple jamming policies since they focus on determining  free channels in future time slots instead of  selecting a countermeasure for  a specific type of jamming attack. 
However, the exploitation of the obtained knowledge is inefficient since these methods only utilize the information  of the spectrum in the last time slot while the information of the spectrum occupancy in previous time slots can be utilized to better understand  the spectrum occupancy pattern.

In light of the above discussion, it is obvious  that the problem of a wireless network under jamming attacks from multiple jammers with multiple jamming policies needs to be investigated. Thus, in the second part of the paper, we propose an anti-jamming technique for multiple jammers with different jamming policies.

 The main contribution of this paper is to propose two novel anti-jamming methods against a single jammer capable of launching attacks with different jamming policies and multiple jammers with different jamming policies. 
In order to defend  the single jammer, we develop an anti-jamming method that requires a small data set for training and is practical against several jamming types. In this context, the jammer's type is first detected, and then an appropriate countermeasure is selected. For  multiple jammers with different jamming policies, we propose to estimate the jammers' future behavior from their occupied channels in previous time slots. In both of the proposed anti-jamming techniques, the occupied channels of the jammers in previous time slots are stored, and based on the obtained information, the jamming policy or jammers' future occupied channels will be estimated. Due to the sequential nature of the interaction between users and  jammers, we propose the use of recurrent neural networks (RNNs) in both cases. To evaluate our methods, we calculate the ergodic rate (ER) and  perform different simulations considering different jamming policies. Our results show that in the single jammer scenario, the jamming policies are detected perfectly with high accuracy within a short period, and as a result, the best countermeasure is employed, leading to a high successful transmission rate  (STR)  and ER  for the user. Moreover, against  multiple jammers, the proposed anti-jamming technique achieves a STR and ER higher than $75\%$ when $70 \%$ of the spectrum is being jammed. This   value rises to $90\%$ when $30 \%$ of the spectrum is being jammed.

\textcolor{black}{  The rest of this paper is organized as follows.  Section  II   presents   the system model and problem formulation. In sections IV  and V, we present the proposed anti-jamming methods for single  and multiple jammer  scenarios, respectively.  Simulation results are
shown in Section VI and  conclusions are drawn in
Section VII.
}

\begin{figure}[!tbp]
\centering
 \includegraphics[width=0.5\textwidth, height=0.3\textwidth]{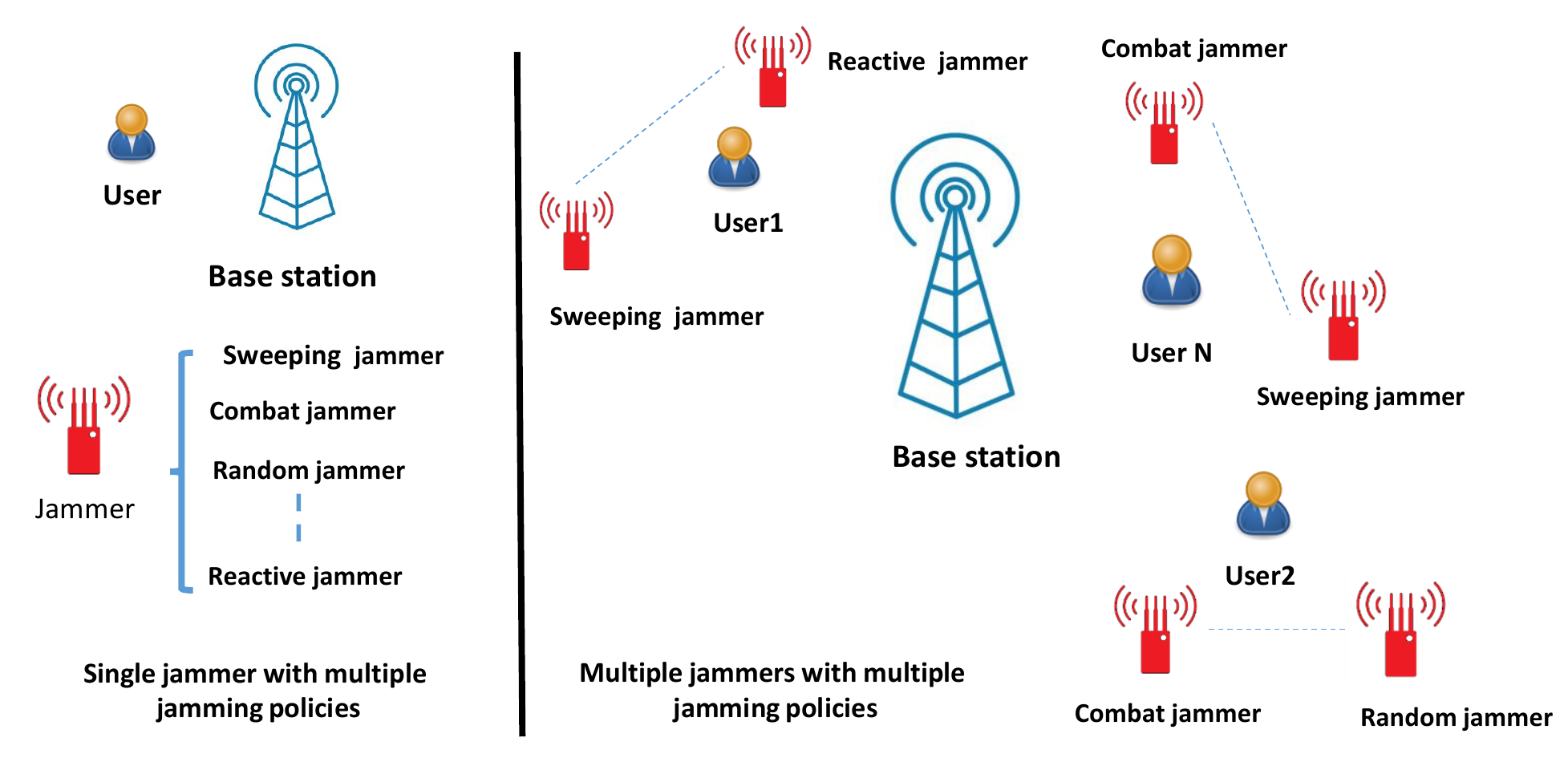}
\caption{Considered system model. }
\label{System_m}
  \end{figure}  
\section{System Model and Problem Formulation}
 
 \subsection{System model}
We consider  a  wireless network composed of  a BS,   $N$ users, and $J$ jammers. We study two
distinct scenarios based on the number of jammers in the
network. In the first scenario (SC1), the network includes  a  user and a jammer capable of attacking with various jamming policies, while in the second scenario (SC2), we assume that multiple users are attacked by multiple jammers with different jamming policies. In  both scenarios, the BS is located at the center of the  network while  users and jammers are randomly distributed in the network,  as shown in Fig. \ref{System_m}. In both scenarios, we assume that the network is constantly under jamming attacks.    In  SC2, each user is   assumed to be located within the communication range of at least one jammer. Time is divided into time slots, where  at each time-slot,  each user is served using a channel selected from $L$   channels. 
  The   packet  transmitted by a user over a frequency channel is successfully received when that channel is not jammed or interfered by other users. 
Four types of   jamming policies are considered, where in each time slot, each jammer can jam several frequency channels.  The considered  jamming policies are:  
\begin{itemize}
\item \emph{Random jammer}, which chooses its channels randomly.

\item \emph{Sweeping jammer}, which periodically shifts its full energy over multiple frequency channels.
\item \emph{Reactive jammer}, which  listens to channels repeatedly and jams channels    after a single time slot from sensing an activity.

\item\emph{Combat jammer}, which  chooses a  number of channels randomly and jams them for a number of  consecutive time slots.
   
\end{itemize}

In SC1, the jammer selects a policy from the above-mentioned policies while in  SC2, we assume that each user is attacked by a group of jammers employing all of the above-mentioned jamming policies. In both scenarios, we assume that each user can sense the jamming signals in the frequency channels to detect occupancy. In this context, if the sensed amplitude of the base-band signal  of a channel is higher than a specific threshold, the channel is counted as an occupied channel,  otherwise, it is assumed to be free.

 \subsection{Problem formulation}
We now formally  formulate the interaction between users and the jammers as an optimization problem. To this end, assuming that  the vector of the indices of all  frequency channels allocated to  users under  policy  $\pi$  is given by
 
\begin{equation}
\boldsymbol{c}^{\pi} = [c^{\pi}_1, c^{\pi}_2, ...,c^{\pi}_N],
\end{equation}
where $c^{\pi}_N$ denotes the index of the $i^{th}$ user's selected channel  under policy  $\pi$, the  instantaneous sum rate of the network at time slot $t$ is defined as 
\begin{equation}
R = \sum_{k=1}^{N}\mathbb{I}(c^{\pi}_k)\log_2\left(1+\frac{\Omega_k|h_{kc_k}|^2}{ \delta^2}\right),
 \end{equation}
where $t$ denotes the time slot index, $h_{kc_k}$ is the channel gain between the $k^{th}$ user and the BS in frequency channel $c_k$, $\Omega_k$ is the $k^{th}$ user's power, $\delta^2$ is the noise power, and  $\mathbb{I}(\cdot)$  is an indicator function  given by 

\begin{equation}
\mathbb{I}(c^{\pi}_k) = \left\{
\begin{array}{ll}
 0, & \mbox{if } c^{\pi}_k \mbox{ is jamming and collision free.}  \\
 1, & \mbox{otherwise,} 
\end{array}
\right.
\end{equation}
  Our goal is to obtain a channel allocation strategy $\pi$ that allocates free frequency channels to the users at each time slot. Hence, we set the instantaneous sum rate of the network as the objective function of the optimization problem. As a result, the design of the optimal strategy $\pi^{*}$ for  the users' channel allocation can be formulated as the following optimization problem:

\vspace{-.25cm}
\begin{equation}
\begin{aligned}
\label{opt}
&   \underset{\pi  }{\textnormal{maximize}  } {\hspace{0.5cm}R}, \\
&\textnormal{s.t.}  \hspace{0.5cm}
    \textnormal{ } |\mathbb{A}| \geq{ J} 
  \textnormal{  and } |\boldsymbol{c}^\pi| = N \hspace{0.1cm} \forall t,
\end{aligned}
\end{equation} 
 where 
$\mathbb{A}=\{\boldsymbol{u}_1, \boldsymbol{u}_2, ...,\boldsymbol{u}_J\} $ denotes the selected frequency channels by the jammers and 
 $\boldsymbol{u}_i$ is a vector containing the  indices vector of the  $i^{th}$ jammer's selected frequencies. 
The constraints $|\mathbb{A}| \geq{ J}$ and $|\boldsymbol{c}^\pi| = N$  imply that each jammer jams at least one channel and one frequency channel is assigned to each user at each time slot, respectively.

In each iteration of the interaction between the users and the jammers, the users must find the optimal channels. Problem (\ref{opt}) is a discrete optimization problem aiming to maximize the users rate by selecting the optimal channel between the users and the BS. In this context, the optimal choice for each user is  to select  the channel  that has the highest channel gain between the user and the BS among free channels. 
In order to characterize the maximum achievable ER, we consider
an ideal case each user knows the jammed channels in
the next time slot. We use a Nakagami $m$-fading distribution with   average channel power
gain  $\lambda$  to model the channel gain between each user and the BS. The total number of jammed channels around the $k^{th}$ user is $\boldsymbol{\upsilon}_k$, the ER of the $k^{th}$ user is obtained using Proposition 1.  

\begin{figure}[!tbp]
\centering
 \includegraphics[width=0.4\textwidth, height=0.4\textwidth]{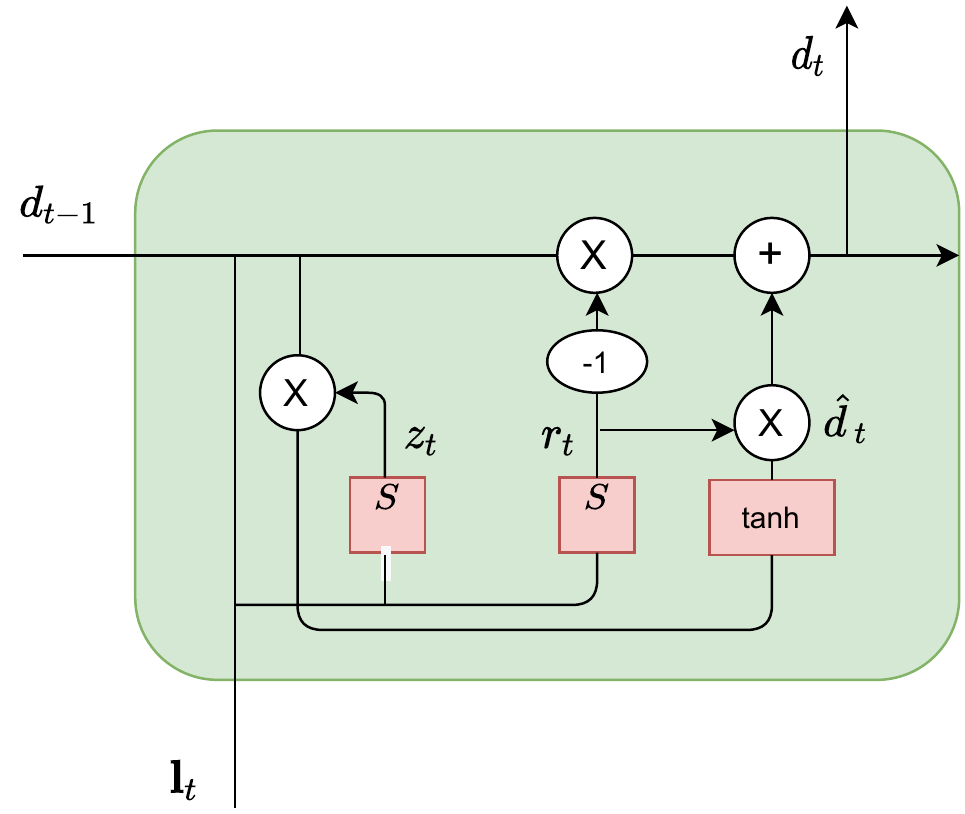}
\caption{GRU network. }
\label{GRUnet}
  \end{figure} 

\begin{proposition} 
\label{pro1}
 \textnormal{The ER of the   $k^{th}$ user when the highest channel gain between this user and the BS is selected will be:}
 \begin{equation}
\begin{aligned}
\label{erg1}
&R_{ek}=\frac{\Omega_k}{ \delta^2\ln{(2)}}\int_{0}^{\infty} \frac{1-(\frac{\gamma(m,\frac{m x}{\lambda})}{\Gamma(m)})^{(L-\upsilon_k)}}{1+\frac{\Omega_kx}{\delta^2}} dx
\end{aligned}
\end{equation}
\begin{proof}
\textnormal{See Appendix A. }
\end{proof}
\end{proposition}

When the users interfere with each other, the ER of each user  can be obtained by slightly modifying (\ref{erg1}). In this context,  the number of free channels for each user reduces to $L-N-\upsilon_k+1$. Thus,  the ER of the $k^{th}$ user   will be:
 \begin{equation}
\begin{aligned}
\label{erg2}
&R_{e\zeta_1}=\frac{\Omega_k }{\delta^2 \ln{(2)}}\int_{0}^{\infty} \frac{1-(\frac{\gamma(m,\frac{m x}{\lambda})}{\Gamma(m)})^{(L-N-1-\upsilon_k)}}{1+\frac{\Omega_kx}{\delta^2}} dx.
\end{aligned}
\end{equation}
In some cases, it is recommended to select the users' channels by following a random policy that can  minimize the probability of being traced by the jammers. In this context, the ER of the $k^{th}$ user  will be obtained using Proposition 2.

\begin{proposition} 
\label{pro2}
 \textnormal{The ER of the $k^{th}$ user when the user selects its channels randomly is given as}
 \begin{equation}
\begin{aligned}
\label{erg3}
R'_{ek}= \frac{1 }{ \Gamma({m})\ln({2})}G_{2,3}^{3,1} \left( \frac{\delta^2 m}{ \Omega_k\lambda }\bigg| \begin{matrix}0, 1\\ 0, 0, m \end{matrix} \right)
\end{aligned}
\end{equation}
\begin{proof}
\textnormal{See Appendix B.}
\end{proof}
\end{proposition}
The equation in (\ref{erg3}) is valid for both the ideal and interference scenarios. For all the aforementioned  scenarios, the ergodic sum rate can  be obtained  by taking the sum of  all the users' rates. We note that the obtained ER  using (\ref{erg1}) and (\ref{erg3})  are the maximum achievable ERs since they are calculated assuming that each user knows the  other users' and jammers' occupied channels in future time slots.
 
 In order to solve (\ref{opt}), the jammers' selected channels  at the next time slot, i.e, $\mathbb{A}$,
 must be known; however, in
realistic scenarios, this information is not available. Therefore, next, we propose RNN-based  anti-jamming methods to predict  the  occupied channels by the jammers in the next time slot. 
\section{Proposed anti-jamming method against single-jammer  }

     \begin{figure}[t]
\centering
 \includegraphics[height=0.5\textwidth,width=0.5\textwidth]{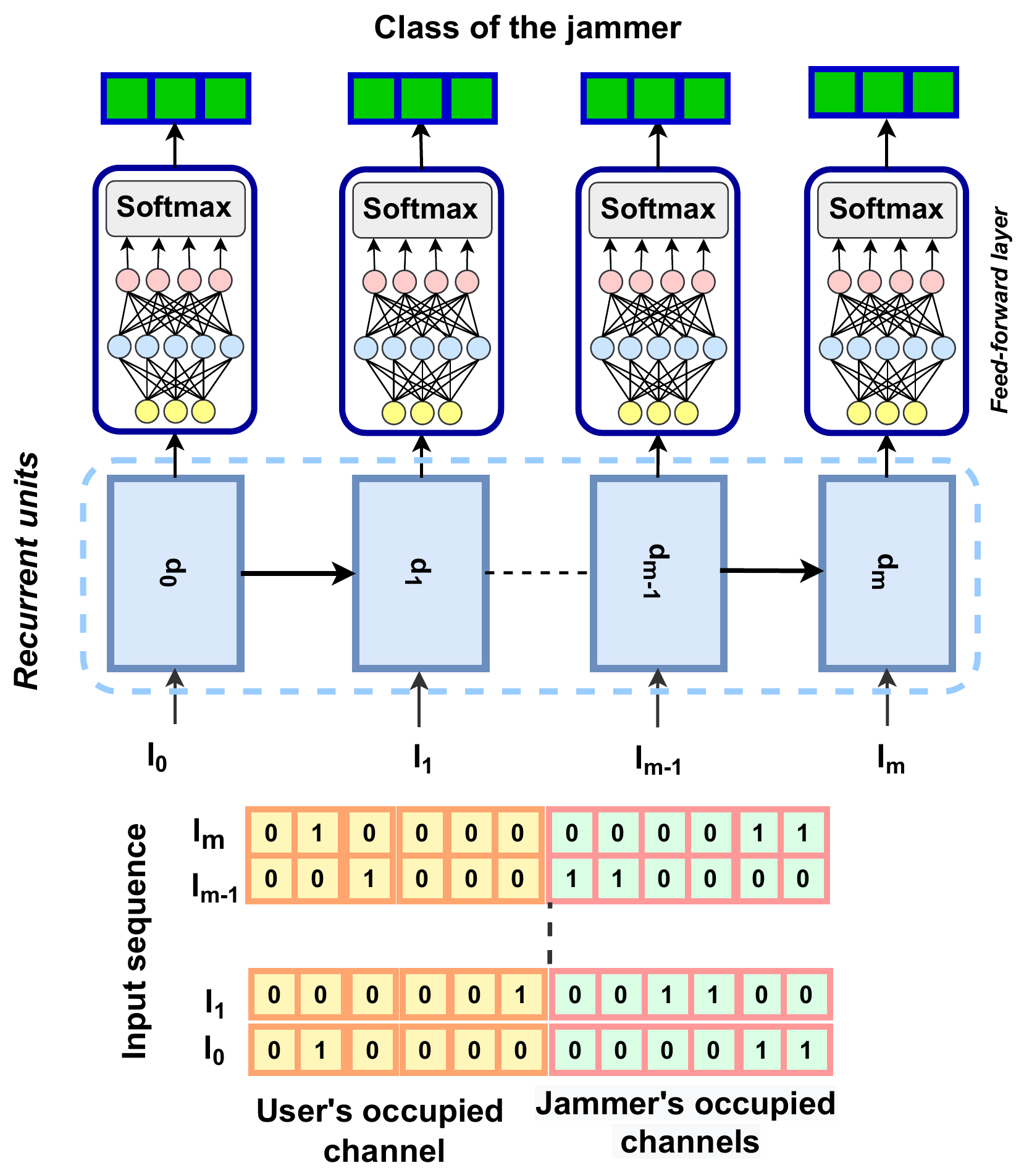}
\caption{Proposed RNN for single jammer scenarios. }
\label{GRU_P1}
  \end{figure}
 
We now introduce the proposed anti-jamming method  in SC1. Since the user  is attacked by  a jammer  that can jam     with different jamming policies,  we propose to first recognize the jamming policy, and then select an appropriate countermeasure against the jammer. In this context, the interaction between   the user  and the jammer is sequential. We thus employ  RNNs because of their ability to process data with a sequential nature.   
RNNs are  artificial neural networks that can learn patterns and long-term relationships from time series and sequential data. At each time slot, an RNN takes the previous hidden state and the input, and   generates   the updated hidden state   and output  as  follows \cite{RNN}
 
  \vspace{-0.6cm}
\begin{equation}
\begin{aligned}
\\&\boldsymbol{d}_t =\sigma(\boldsymbol{K}_d \boldsymbol{d}_{t-1}+\boldsymbol{V}_{\boldsymbol{d}}\boldsymbol{l}_t +\boldsymbol{g}_{\boldsymbol{d}})
\\&\boldsymbol{y}_t =\sigma(\boldsymbol{K}_y\boldsymbol{d}_t +\boldsymbol{g}_{\boldsymbol{y}}),
\end{aligned}
\end{equation}
where  $\boldsymbol{l}$ is the input,  $\boldsymbol{d}_t$ is the hidden state at time slot $t$,  $\sigma$ is an activation function, $\boldsymbol{V}_{\boldsymbol{d}}$ is the weight of the input vector, $\boldsymbol{g}_{\boldsymbol{d}}$ and  $\boldsymbol{g}_y$ are   the bias terms, and $\boldsymbol{K}_{\boldsymbol{d}}$ and  $\boldsymbol{K}_y$ denote the weights of the hidden layer and the output, respectively. High depth and recurrent connections 
in conventional RNNs cause a vanishing gradients problem. This vanishing gradient challenge  is  addressed  in   gated recurrent unit (GRU)\cite{chung2014empirical} models by  controlling the inputs  using multiple gates in a hidden layer.

In this context, based on the conditions of the input and previous hidden layer, the  controller gate $r_t$ is  set to one or zero to update the hidden layer. The overall process is  shown in Fig. \ref{GRUnet}. At time slot $t$,  the   GRU states are updated as follows:
 \vspace{-0.7cm}
 
\begin{equation}
\begin{aligned}
\\&\boldsymbol{r}_t =S(\boldsymbol{W}_{r}\boldsymbol{l}_t +\boldsymbol{U}_{z}\boldsymbol{d}_{t-1}+\boldsymbol{g}_r)\\&
\boldsymbol{z}_t = S(\boldsymbol{W}_{z}\boldsymbol{l}_t +\boldsymbol{U}_{z}\boldsymbol{d}_{t-1}+\boldsymbol{g}_z) 
\\&\hat{\boldsymbol{d}_t} = \tanh(\boldsymbol{W}_{{\boldsymbol{d}}}.\boldsymbol{l}_t +\boldsymbol{U}_{\boldsymbol{d}}.(\boldsymbol{z}_t  \odot \boldsymbol{d}_{t-1})+\boldsymbol{g}_{ {\boldsymbol{d}}})
\\&\boldsymbol{d}_t = (1-\boldsymbol{r}_t) \odot \boldsymbol{d}_{t-1} +\boldsymbol{r}_t \odot \hat{\boldsymbol{d}_t},
\end{aligned}
\end{equation}
where $\boldsymbol{W}_{\boldsymbol{r}}$, $\boldsymbol{U}_{\boldsymbol{r}}$, $\boldsymbol{W}_{l\boldsymbol{z}}$, $\boldsymbol{W}_{\boldsymbol{z}}$, $\boldsymbol{W}_{\boldsymbol{d}}$,    $\boldsymbol{U}_{\boldsymbol{d}}$, $\boldsymbol{g}_{\boldsymbol{r}}$, $\boldsymbol{g}_{\boldsymbol{z}}$, and  $\boldsymbol{g}_{\boldsymbol{d}}$  are learning parameter   matrices and vectors. Here the operator   $\odot$ represents the Hadamard product, and  $S$ is the Sigmoid function.
      \begin{figure}[t]
\centering
 \includegraphics[height=0.5\textwidth,width=0.5\textwidth]{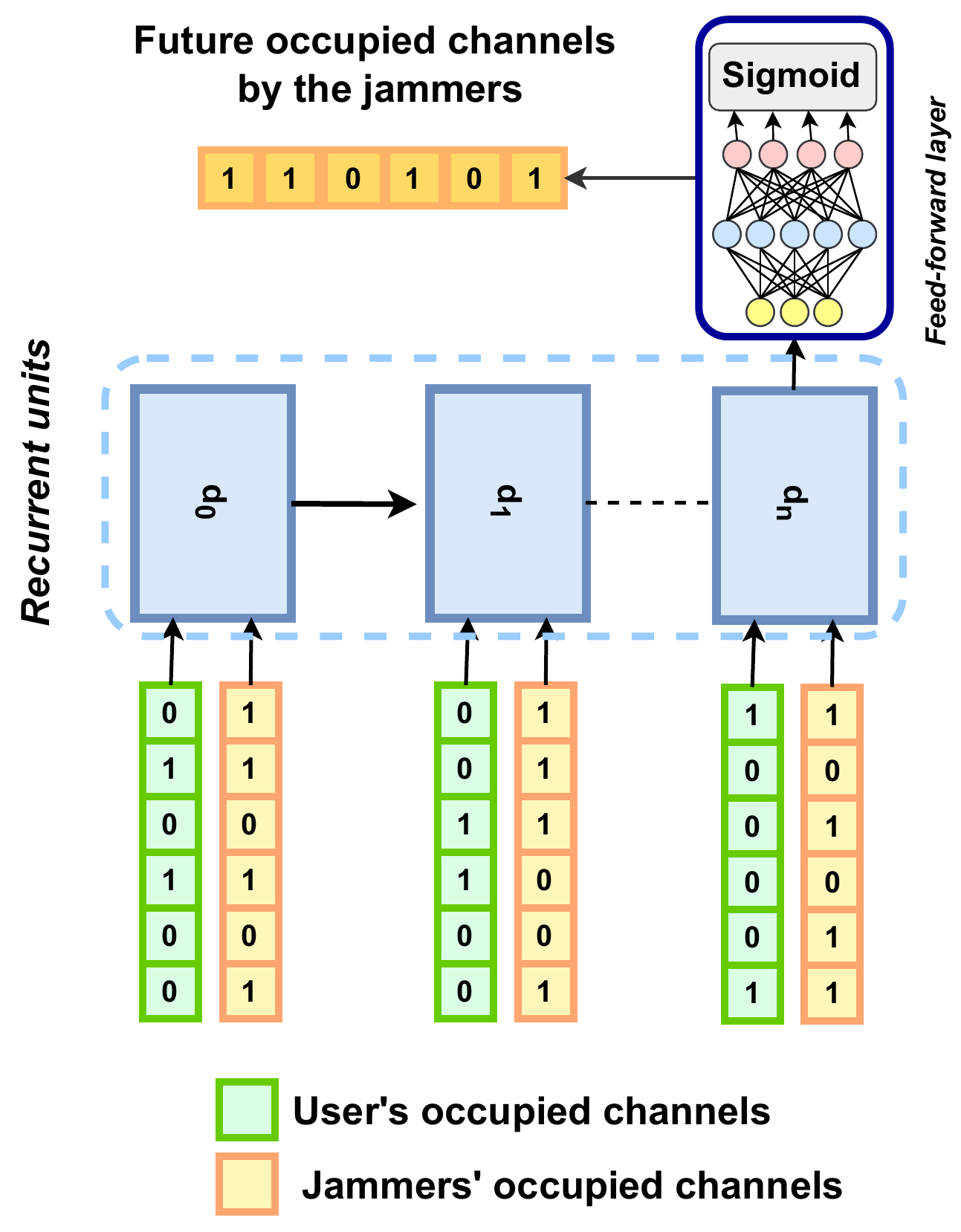}
\caption{Proposed RNN  for multiple jammers  scenarios. }
\label{GRU_P2}
  \end{figure}

 A feed-forward layer with an output size equal to the number of considered jamming classes is considered after each GRU hidden state. Then, the output of  the feed-forward layer  is passed through a SoftMax layer to generate a probability distribution vector for the jamming type classes.  It is noted that the predicted class is obtained by the index of the output array that has the highest probability value.
 In Fig. \ref{GRU_P1}, we show the proposed  network.

We divide the proposed recognition technique into training and testing. In the training phase, which is an offline process, the interaction between  the  user and the jammer is simulated. In contrast, the testing is an online process that takes place during the  interaction   between the user  and the jammer.

The interaction between the user and the jammer is  simulated  for $T$ consecutive time slots. During the simulation, a channel is randomly assigned to the user and the jammer's responses  for all of  the considered jamming policies, i.e, random, sweeping, reactive, and combat, are simulated  and the  channels of the users, the jammer, and  corresponding jamming policies are saved. At each time slot, a  vector with $2L+1$   elements is considered.
The first  and second $L$ elements are  used to save the user's and jammer's utilized channels, respectively. Specifically, the $i^{th}$ element of the vector   is set to one if the $i^{th}$ channel   is occupied by the user, otherwise it is set to zero.  The same procedure is used  to denote
the jammer’s occupied channels.
Finally,  the jamming policy class is stored in  the $(2L+1)^{th}$ element. Given that the number of channels jammed in each time slot, in addition to the jammers' channel selection policies, is crucial, a distinct class is assigned to each combination of jamming policy and number of jammed channels. For instance, a sweeping jammer that jams two channels per time slot is not in the same class as one  that jams one channel. As a result, a sweeping jammer with one to four channels must have four classes assigned to it.  

The collected data  from the simulations is then used to train the proposed RNN. Here, the elements with column indices from $1$ to $2L$ are  the inputs while elements in the $2L+1^{th}$  column are the corresponding    class targets. During every step of the training, a matrix consisting of $a$ consecutive vectors is fed to the RNN and  a vector  with the same number of elements as the number of classes is generated  by the network. We employ the cross-entropy  as the loss function to train the network.

  In the testing phase,  similar to the training process, the user's and  jammer's  occupied channels in the last $b$ time slots are saved. In this context,  the selected channels by the user  and jammer during the last $b$ time-slots are  given to the trained network, and the network then generates a vector.   The class of the jammer is determined by  the  index of  the highest value in the output vector.
Anti-jamming can be easily selected once the jamming policy and corresponding number of jammed channels are identified. In this context, the future behavior of the  combat, sweeping, and reactive  jammers can be predicted from the  detected class of the jammer and  the jammer's  selected channels in previous time slots. Thus, the user  can find the best   channel  that maximizes its rate among free channels. For instance, if the trained network detects that the channels are jammed by a sweeping jammer in which jams three channels per time slot,  users predict that the  sweeping jammer will jam channels with indices $5$ to $7$ in the coming time slot  when the indices of jammed channels in previous time slot are $2$ to $4$. Thus, users will select  channels other than channels with indices $5$ to $7$. For the random jammers, the future channels that the jammer will select cannot be determined. In this case, the user  selects a channel randomly among all channels. 
     
 




\section{Proposed anti-jamming method against multiple jammers  }
In SC2, users are simultaneously attacked by  multiple  jammers with different jamming policies. In this context, users cannot differentiate between the jamming policies of each jammed channel, and thus the proposed anti-jamming method in SC1  cannot be employed. We propose to estimate the channels that will be jammed  in the next time slot  
using the jammers' occupied channels in previous time slots. Once more, the interaction between  the users and the jammers has  a sequential nature, and therefore  we also use RNNs in SC2. We consider two scenarios depending on whether or not users interfere with  each other. Since we assume that the status of a frequency channel is solely determined  by sensing  the signal amplitude  of that channel and   no signal processing is performed, the users cannot distinguish between a jammed and an interfered channel.

In SC2, we consider an   RNN for each user.  In this context, after each   GRU hidden layer,  we use a feed-forward layer with output size equal to the number of channels, as depicted in  Fig. \ref{GRU_P2}. Next, the output of the fully connected layer is fed into a Sigmoid function to normalize the output  elements between zero and one. Afterwards, the values that are  higher than $0.5$ are set to one and the ones lower than $0.5$ to zero.

 Similarly, the proposed anti-jamming method is  divided into two phases, i.e.  training and testing.  In each time slot of the training phase, each user  considers a vector with $2L$ elements, and   saves its  utilized channel  in the first $L$ elements  while  the occupied channels by jammers  are saved in the second $L$ elements (and other users in case of interference). Particularly, the $i^{th}$ element   is set to one if the $i^{th}$ channel   is occupied, otherwise it is set to zero. During training,  we use a  random channel selection policy for the users so as to generalize  the results. However, the proposed training scheme can work with any other channel allocation methods. To provide the target data, the vector containing the occupied channels in the current time slot is used as the target of the previous time slots because we want to estimate the occupied channels in the next time slot.  
 
 During every step of the training, a matrix consisting of $a$ consecutive vectors is fed to the RNN, and  a vector with $L$ elements is  generated  by the network. The binary cross-entropy
is utilized as the loss function to train the network. The testing process of SC2 is similar to  that in  SC1 with a slight change that 
estimated occupied channels in the next time slot is outputted
from the  trained network. 

     \begin{figure}[t]
\centering
 \includegraphics[height=0.45\textwidth,width=0.5\textwidth]{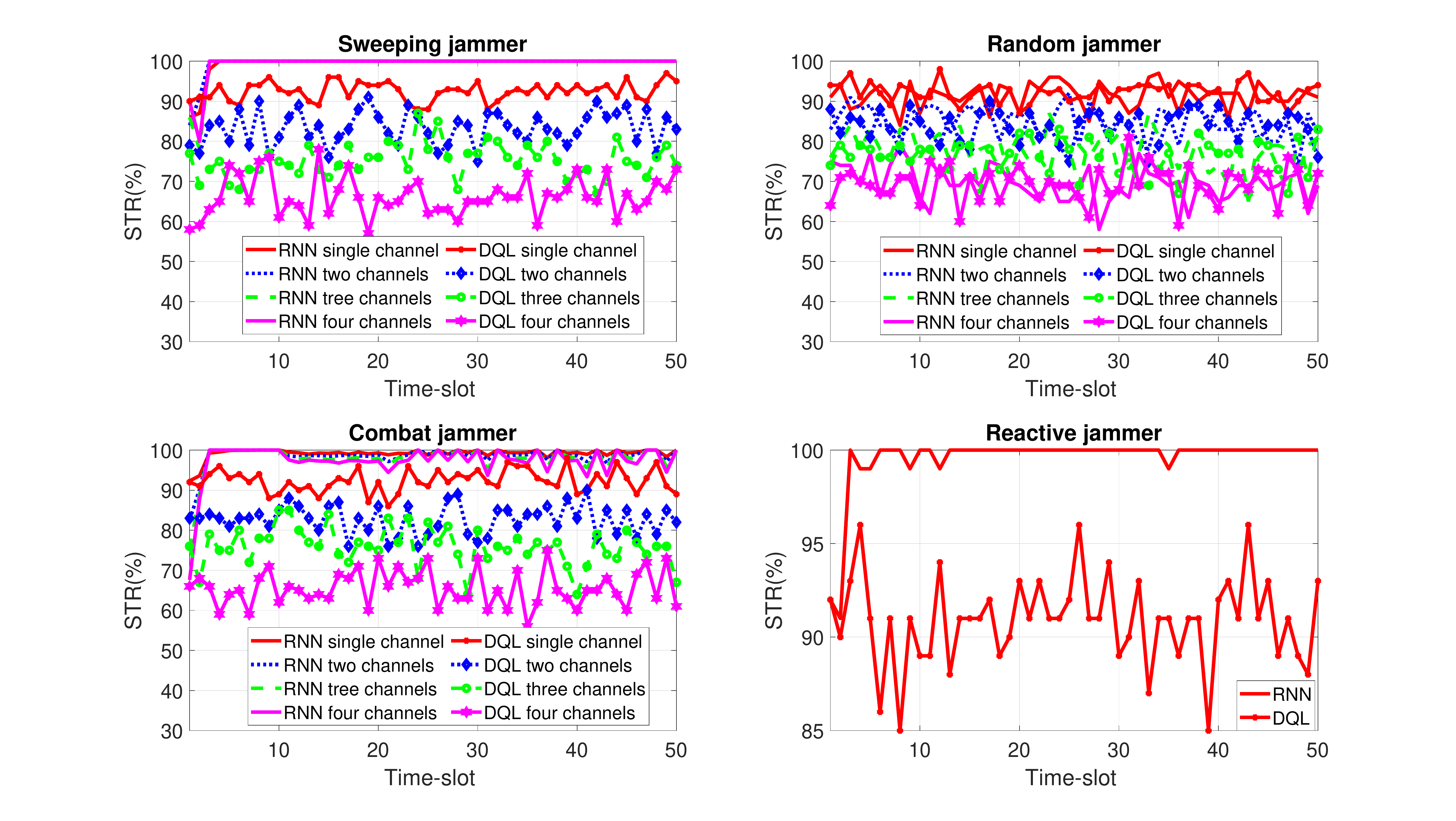}
\caption{STR of the proposed anti-jamming method. }
\label{fig2}
  \end{figure}

In the two anti-jamming techniques proposed in this work, the information about the jammed channels  is required. To extract this information, each user senses all the frequency channels and flags channels where the amplitude of the base-band signal is higher than a specific threshold. In this context, the received jamming signal in a frequency channel is given as:
\vspace{-0.1cm}
 \begin{equation}
\chi_{k} = \Upsilon_{k}+n,
\end{equation}
 where $\Upsilon_{k}$ is the amplitude of the jamming signal at the $k^{th}$ user's side and $n$ is  the white Gaussian noise with  power  $\delta^2$.
 Assuming that the detection  threshold  is $\Gamma$, a miss-detection happens when the noise amplitude is lower than $  n \leq (\Gamma-\Upsilon_{k}) $ and a false alarm happens when $ n \geq  \Gamma    $. Thus, the probability of miss-detection $P_{\textnormal{md}}$ and false alarm $P_{\textnormal{fa}}$    are given as 
 

 \begin{equation}
P_{\textnormal{md}} = Q\left(\frac{ \Upsilon_{k} - \Gamma  }{\delta}\right)   ,
\label{fa}
\end{equation}

  \begin{equation}
P_{\textnormal{fa}} =  Q\left(\frac{  \Gamma }{\delta}\right),
\label{md}
\end{equation} 
respectively.  $P_{\textnormal{fa}}$ shows that the probability of false alarm only depends on the value of the threshold and noise power while $P_{\textnormal{md}}$ depends on the jamming signal power at the users' side. 

 $\Gamma$ must be high enough  to minimize $P_{\textnormal{fa}}$, and should not be too high since increasing $\Gamma$ increases  $P_{\textnormal{md}}$.  The value of $\Gamma$ can be selected based on  the desired values for $P_{\textnormal{md}}$ and  $P_{\textnormal{fa}}$,  and the jamming- to-noise-ratio (JNR). For instance, $\Gamma$ should satisfy $\frac{\Gamma}{\delta}=2.32$   to have $P_{\textnormal{fa}} = 0.01$. Here,  $P_{\textnormal{md}}=0.01$ for a JNR of $\frac{||\Upsilon_k||^2}{\sigma^2} = 13.35$ dB.  For both of the considered scenarios, we set $\frac{\Gamma}{\delta} = 2.8117$ and JNR $=15$ dB.

\section{Simulation Results}
We now assess the  proposed anti-jamming methods in  both considered scenarios using extensive simulations. To this end,  we define an evaluation metric, named STR, which quantifies the ratio of the number of successfully delivered packets  to all the transmitted packets. For the first  scenario, the proposed anti-jamming method is assessed by evaluating the STR and  detection accuracy for  the different  jamming policies  as a function of the elapsed time. We consider  that  random, combat, and sweeping jammers can simultaneously jam one to four channels. Moreover, we compare   the output of the proposed anti-jamming method  to the case in which   the user employs deep Q-learning (DQL) for its anti-jamming and channel allocation policy.  In addition, we compare the  ER obtained by the proposed anti-jamming techniques to the calculated   ER in (\ref{erg1}) and (\ref{erg3}). In SC1, simulation results are obtained using $12$ channels. For SC2, we assume that each user is surrounded by a random, a combat, a  sweeping, and a reactive jammer. The jammers' specifications  are provided in Table \ref{table:1}. The proposed anti-jamming method is evaluated by illustrating the STR  of one to four  users when  $30\%$  to $70\%$    of the spectrum is under jamming attacks. Moreover, results are compared to the case where DQL is employed for the users' channel allocation. In addition,  we show the performance of each jammer in terms of successful jamming attacks to clarify the impact of each jamming type separately. In SC2, simulation results are obtained by considering $20$ channels. Moreover, we choose $a$ and $b$ as $20$, and channels are assumed to follow Rayleigh fading, i.e. $m=1$.  
   
\begin{table}[t!]
\centering
\caption{Number of jammed channels by jammers.}
\label{table:1}
\begin{tabular}[th]{|c|c|c|c|c|c|c|}
  \hline
 &\multicolumn{5}{| c |}{Jamming ratio}    \\
 \hline
 Jammer type     &30\% & 40 \% & 50\% & 60\% & 70\% \\
\hline
Sweeping jammer&2 & 3 & 4 & 4  & 5 \\ \hline
Reactive jammer& 1 & 1 & 1 & 1  & 1  \\ \hline
Random jammer &1 & 2 & 2 & 3  & 3  \\ 
\hline
Combat jammer&2& 2 & 3 & 4  & 5\\ \hline
\end{tabular}
\end{table} 


\subsection{Anti-jamming in SC1 scenario}

In Fig. \ref{fig2}, we show the STR resulting from the proposed  anti-jamming method for all the considered types of the jammers. From this figure, we observe   that after  a few   time slots, except for the random case, the obtained STR surpasses  $90\%$   and STRs of almost $100\%$  are obtained  after $20$ time slots. The STR  increases with time since the users are provided with more information from the jammers, which leads to jamming policy detection  with a higher accuracy. As a result of a precise jamming policy detection, an appropriate countermeasure against the detected jammer is taken, resulting in a high STR. Lower STR is obtained in the context of  random jammers  since future channels  cannot be estimated even for perfect jamming policy detection. As a result, users have to select their future channels randomly. In this context, increasing the number of jammed channels by the random  jammer  increases  the probability of  getting jammed by the jammer.  In addition, Fig. \ref{fig2}  shows that the proposed anti-jamming method outperforms the DQL-based anti-jamming method for all the considered cases, except for the Random jammer. Specifically ,  the proposed method converges  within a few time slots while DQL needs more time slots to train the users.

\begin{figure}[t]
\centering
 \includegraphics[height=0.45\textwidth,width=0.5\textwidth]{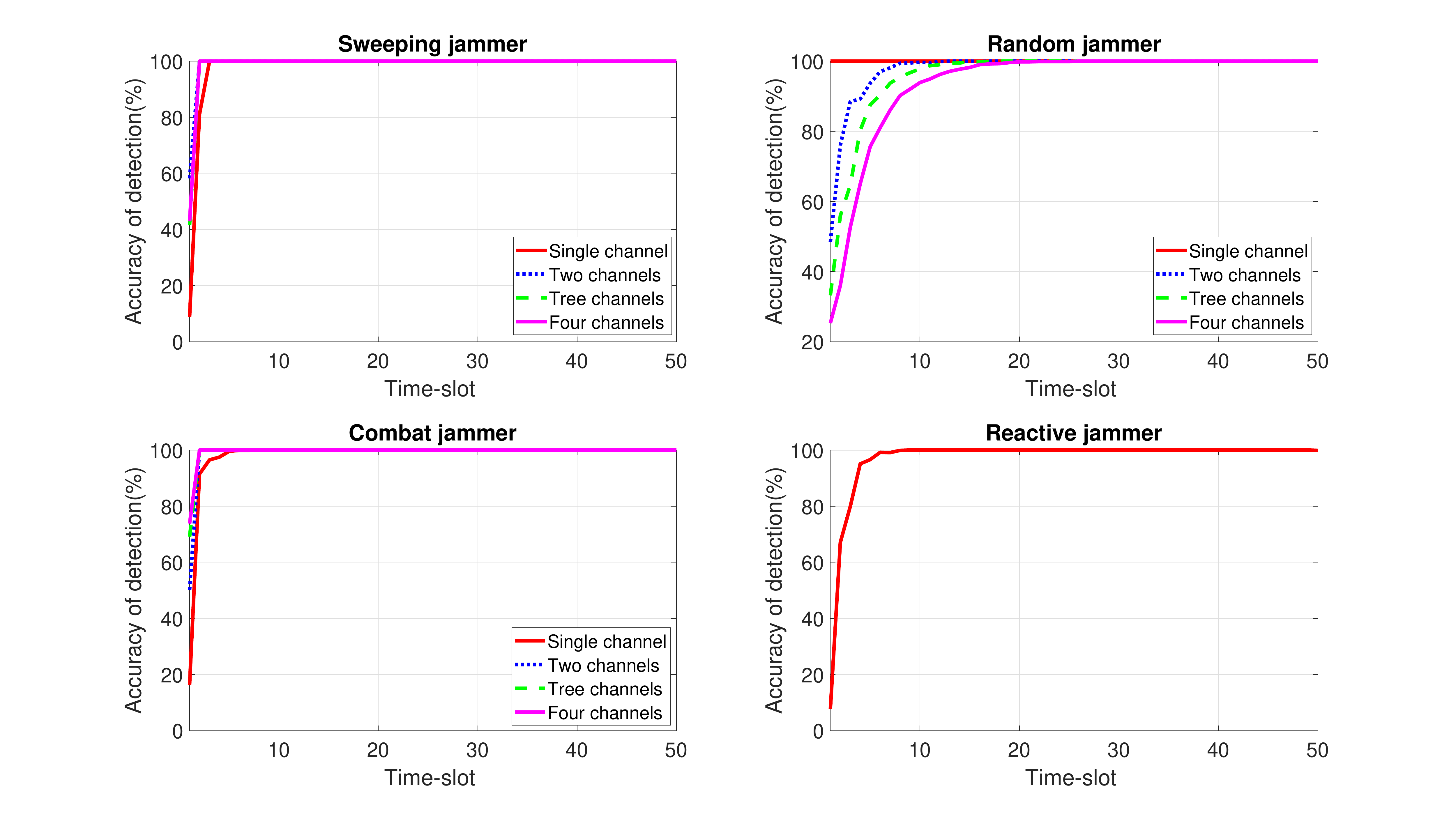}
\caption{ Detection accuracy of the jamming policies as a function of the elapsed time. }
\label{Per_p1}
  \end{figure}

Fig. \ref{Per_p1} presents the detection accuracy rate  for all the considered jamming policies.  From this figure, we observe that, after just five   time slots, the sweeping and combat jamming polices are detected perfectly. Detecting  the   policy of the random and reactive jammers requires more time slots due to the  similarity between these jammers' behavior    in initial time slots.

\begin{figure}[t]
\centering
 \includegraphics[height=0.45\textwidth,width=0.5\textwidth]{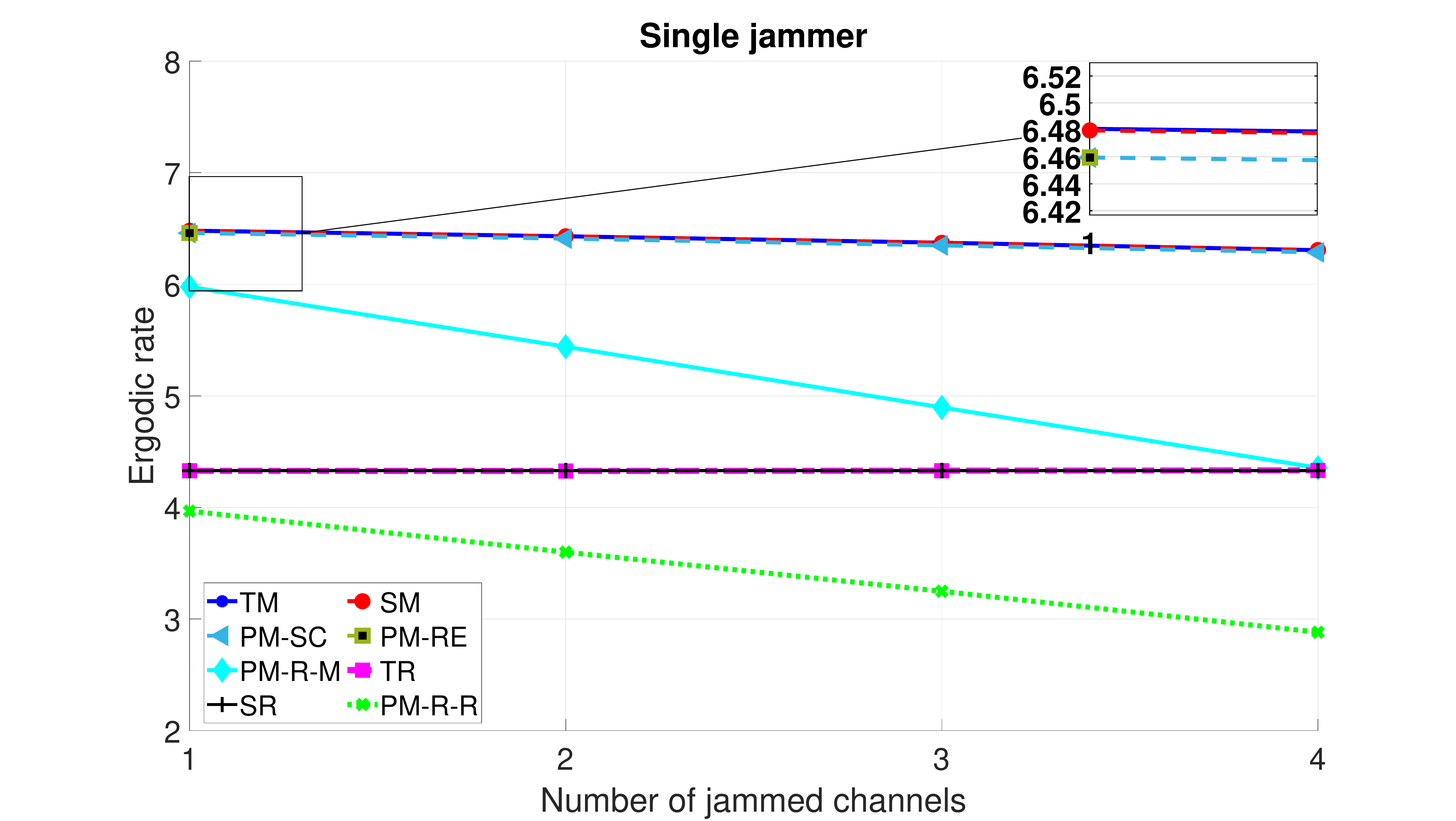}
\caption{ER for SC1. }
\label{erg1f}
  \end{figure}

In Fig. \ref{erg1f}, we verify the ERs derived is (\ref{erg1}) and (\ref{erg3}) using simulations. Here, we consider the number of jammed channels to vary between  one to four. The obtained ERs by the proposed anti-jamming method for all the considered jamming scenarios are shown. The The presented graphs  in order are: 1) ERs obtained by (\ref{erg1}) with legend of  \emph{TM}, 2) simulation to verify (\ref{erg1}) with legend of  \emph{SM}, 3)  proposed anti-jamming method against sweeping and combat jammers with legend of  \emph{PM-SC}, 4) ERs obtained by (\ref{erg3}) with legend of \emph{TR}, 5) simulation to verify (\ref{erg3}) with legend of \emph{SR}, 6) proposed anti-jamming method against random jammer by assigning the  channel with the highest channel gain to the user with legend of \emph{PM-R-M}, and 7) proposed anti-jamming method against random jammer with random channel selection  with legend of \emph{PM-R-R}. Moreover, in Fig. \ref{erg1f}, we present the ER resulting from proposed anti-jamming method when the number of jammed channels is one by the legend of \emph{PM-RE}.

The results in Fig. \ref{erg1f} demonstrate  that the calculated ERs  match   perfectly with the simulation results, which proves the accuracy of the derivations. From this figure, we can also see that the  ERs in the presence of   reactive, sweeping, and combat jammers are close to the maximum achievable ERs.  This, in turn, shows     the effectiveness  of the proposed anti-jamming method in recognizing the jamming policies and assigning high-quality channels to the user. Against the random jammer,  the ER  is less than the other jammers since the random jammer's occupied channels in the next time slot are not predictable. Moreover, for all the considered numbers of jammed channels by the random jammer, assigning the channel with the highest channel gain to the user provides a higher ER  than the case in which the channel assignment is random. 
\subsection{Anti-jamming in SC2 scenario}

\begin{figure}[t]
\centering
 \includegraphics[height=0.45\textwidth,width=0.5\textwidth]{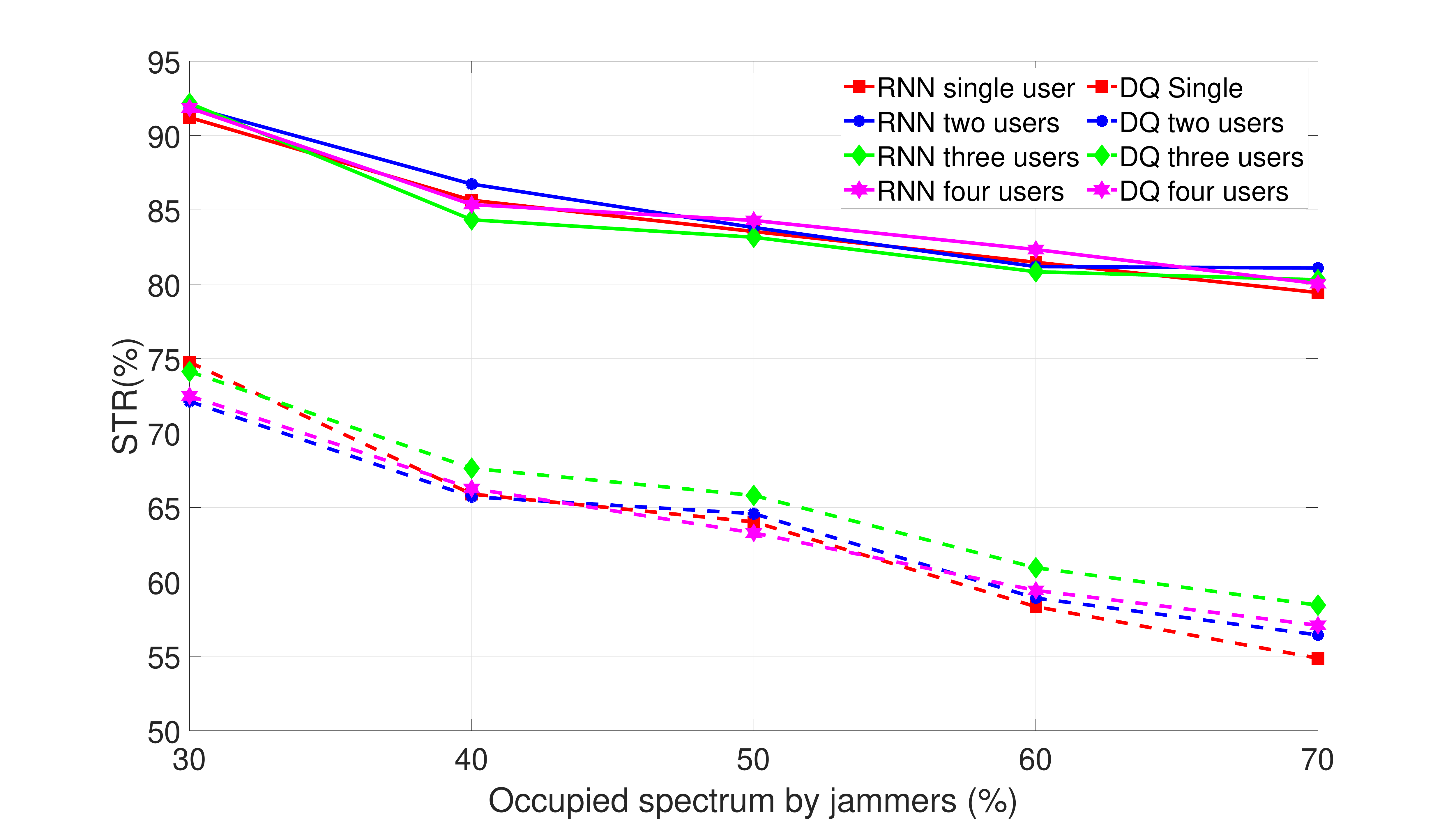}
\caption{STR of the proposed anti-jamming method when users interfere each other. }
\label{per2_p1}
  \end{figure}

  Fig. \ref{per2_p1} presents   the users'  STR   as a function of the elapsed time slots. From Fig. \ref{per2_p1}, we can see that the  users achieve a STR  higher than $90\%$  when $30\%$   of the spectrum is under jamming attacks. This value decreases to   $80\%$    when jammers jam  $70\%$  of the spectrum. The reason behind this performance degradation is that the users have less free channels to utilize. Moreover, for all the considered scenarios, the proposed method outperforms the DQL-based method with a significant gap.

\begin{figure}[t]
\centering
 \includegraphics[height=0.45\textwidth,width=0.5\textwidth]{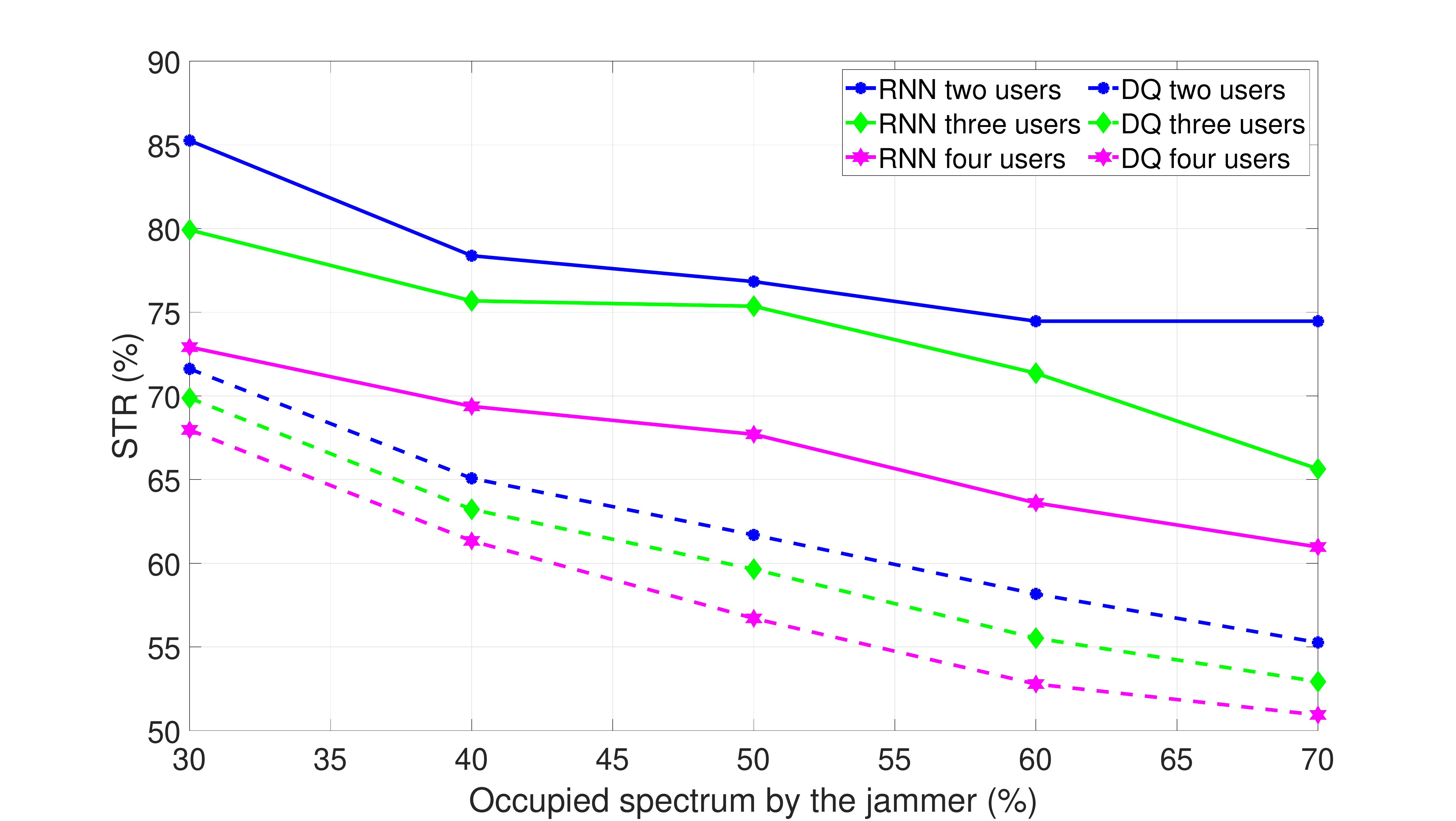}
\caption{STR  of the proposed anti-jamming method when users interfere with each other. }
\label{fig5}
  \end{figure}

The STR  of the  users case of interference is shown in Fig. \ref{fig5}. The performance of the proposed anti-jamming method decreases compared   to the interference-free case. This is because  the RNN is trained assuming that users choose their channels at random. As a result, each user randomly estimates the future behaviour of other users during the testing process. Moreover, when users interfere with each other,  fewer available channels remain  for  channel selection. For instance, in the four users scenario  with $70\%$ of the channels jammed, four users must select their channels   out of six available channels while for the case in which the users do  not interfere, each user  has six available channels.

\begin{figure}[t]
\centering
 \includegraphics[height=0.45\textwidth,width=0.5\textwidth]{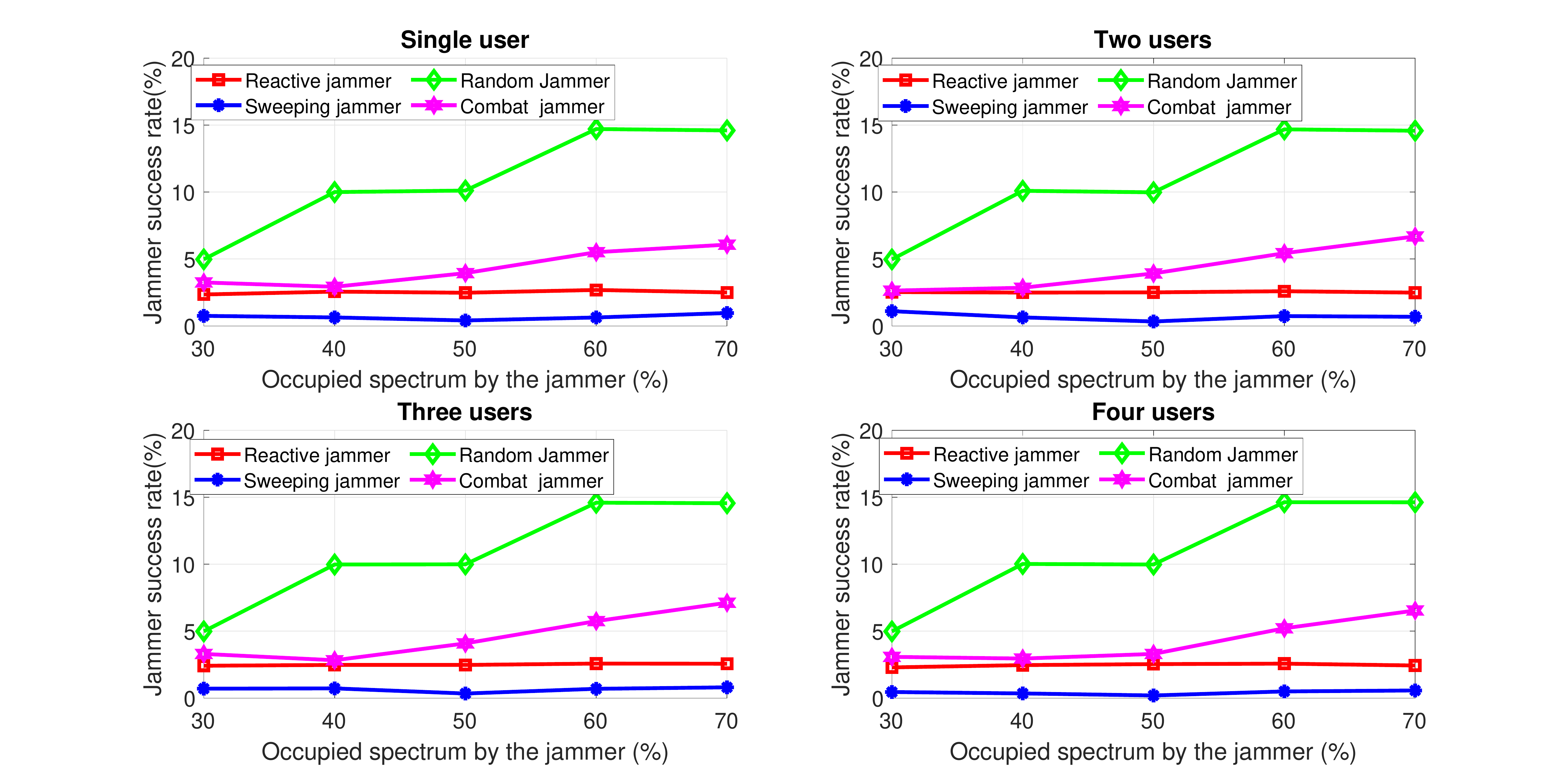}
\caption{Jammers' success rate. }
\label{fig6}
  \end{figure}

\begin{figure}[t]
\centering
 \includegraphics[height=0.45\textwidth,width=0.5\textwidth]{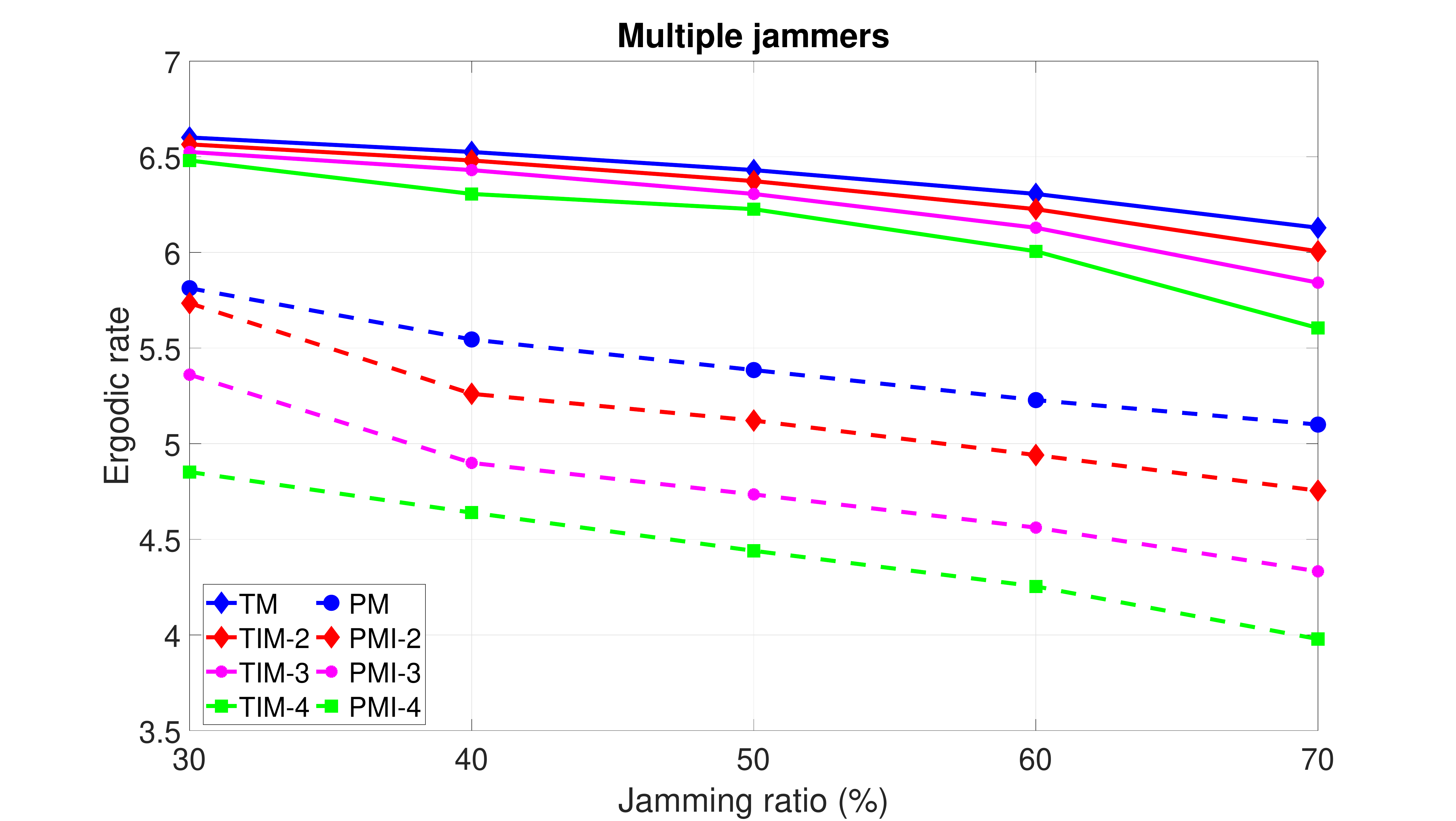}
\caption{ER against  multiple jammers. }
\label{fig611}
  \end{figure}

In Fig. \ref{fig6}, we present each   jammer's  success rate  individually. These  results show that   the   reactive and sweeping jammers have  constant jamming success rates for all the considered  jamming ratios. Meanwhile, the random jammer's success rate increases by increasing the jamming ratio. This is because  the reactive  jammer has a fixed number of jammed  channels for all the considered jamming ratios. Moreover,   the future behaviour of  sweeping jammer is expectable and as a result, increasing the number of jammed channels in this  type of jammer does not impact the success rate. However, it is impossible to predict the channels that the random jammer will choose  in the following time slot. Therefore, increasing the number of jammed channels increases the random jammer's success rate. Similar to random jammers, increasing the number of jammed channels by the combat jammer  increases the success rate of the jammer due to randomness in  the channel selection  of the combat jammer.   However, since the combat jammer jams the selected channels for a number of consecutive
time slots, the increase in the    success rate is lower than that of the random jammers.  
Fig. \ref{fig611} compares between the ERs of
each user obtained by the proposed method, (\ref{erg1}), and (\ref{erg2}), for both cases with and without interference. Drawn graphs in
order are related ERs  obtained by: 1)  (\ref{erg1}), denoted as \emph{TM}, 2) (\ref{erg2}) for two users, denoted as \emph{TIM-2}, 3) (\ref{erg2}) for three users, referred to as \emph{TIM-3}, 4) (\ref{erg2}) for four users, denoted as \emph{TIM-4}, 5) proposed method for the case where users do not interfere each other denoted as \emph{PM}, 6) proposed method for two users scenario while users  interfere each other specified with legend \emph{PMI-2}, 7) proposed method for three users scenario while users  interfere each other, denoted as \emph{PMI-3}, and 8) proposed method for four users scenarios considering  users  interfere each other, denoted as  \emph{PMI-4}. When users do not interfere with each other, the ER of each user is similar over the  different scenarios with different numbers of users. Hence, only the ER  of the single-user scenario is plotted. The    \emph{TM} graph shows that the ER  is higher when there is no interference. Moreover, when users interfere with each other, increasing the number of users and channels decreases the ER since increasing the number of users and jammed channels decreases the number of available channels for each user. The same trend  holds for increasing the number of users and channels in the obtained results by the proposed method, i.e. \emph{PM}, and  \emph{PMI-2} to \emph{PMI-4}. In the case where users do not interfere with each
other, results show that the proposed method achieves ERs  higher than $80\%$  of the maximum achievable ER  for all the considered numbers of jammed channels. Moreover, when $30\%$ of the spectrum is being jammed, the ER  is close to $90\%$ of the maximum achievable ER.
In the case where users interfere with each other, the discrepancy between the maximum achievable ER and the obtained ER by the proposed method is higher than the case of no interference  since fewer free channels are available for the channel selection of each user.

\section{Conclusion}
In this paper, we  proposed  RNN-based anti-jamming techniques against single jammer and multiple jammers. Two
distinct scenarios based on the number of jammers in the
network have been considered. In the first scenario, the network includes a  user and a jammer capable of attacking with various jamming policies, while  in the second scenario, we have assumed that multiple jammers attack multiple users with different jamming policies. Moreover, we have studied two different cases based on 
whether users interfere with each other or not. For both of the considered cases, we have calculated the maximum achievable ER. To evaluate the proposed anti-jamming methods, we have performed extensive simulations assuming four jamming policies. Moreover, we have compared the obtained results with the case where DQL is employed. The results show that against a single jammer, all  the considered jamming policies are detected with high accuracy within a short period, and as a result of an accurate jamming type detection, high  STRs and ERs are obtained. Against multiple jammers, STRs and ERs near $80\%$ are obtained when $70\%$ of the spectrum is under jamming attacks. These values rise to $90\%$ when $30\%$ of the spectrum is under jamming attacks. Moreover, for all the considered  numbers  of users and jamming ratios, the proposed anti-jamming technique outperforms the DQL algorithm with significant gaps.     
  
\appendices
 
\section{Proof of Proposition \ref{pro1}}
\noindent The ER of the $k^{th}$ user is    
\begin{equation}
\label{eq1}
R_{ek} = \mathbb{E}( \mathbb{I}(c^{\pi}_k)\log_2(1+\frac{\Omega_k|h_{kc_k}|^2}{ \delta^2})).
 \end{equation}
Assuming that the jammed channels and occupied channels by other users are known, then $k^{th}$ user selects its channels among free channels, and given $x=|h_{kc_k}|^2$, we can rewrite (\ref{eq1}) as follows:   
\begin{equation}
\begin{aligned}
\label{eq2}
R_{ek} &= \mathbb{E}( \log_2(1+\frac{\Omega_kx}{ \delta^2}))  \\&=\frac{1}{\ln{2}}\int_{0}^{\infty} \ln{(1+\frac{\Omega_k x}{\delta^2})} f_{h_{kc_k}}(x)dx,
 \end{aligned}
 \end{equation}
 where  $f_{h_{kc_k}}(x)$ is the probability distribution function (PDF) of $x$.
 The channel $h_{kc_k}$ is selected among $L-\upsilon$ channels,  where the power gain of each of the channels is a random variable with PDF  
 \begin{equation}
\label{eq3}
f(x)=(\frac{m}{\lambda})^m \frac{x^{m-1}}{\Gamma(m)}
exp[\frac{-mx}{\lambda}],
 \end{equation}
 
\noindent and cumulative distribution function (CDF)  
 \begin{equation}
\label{eq41}
F(x)= \frac{1}{\Gamma(m)}\gamma(m,\frac{mx}{\lambda})=1-\frac{1}{\Gamma(m)}\Gamma(m,\frac{mx}{\lambda}),
 \end{equation}
 where $\gamma(\cdot,\cdot)$ and   $\Gamma(\cdot,\cdot)$ are the  lower incomplete gamma and  upper incomplete gamma functions, respectively.
Since $|h_{kc_k}|^2$ has the highest channel  gain among all the $L-\upsilon$ channels, the CDF of  $|h_{kc_k}|^2$ is given by

 \begin{equation}
\label{eq4}
F_{h_{kc_k}}(x)= \left(\frac{1}{\Gamma(m)}\gamma(m,\frac{mx}{\lambda})\right)^{(L-\upsilon)}.
 \end{equation}

\noindent Integrating by parts, (\ref{eq2}) becomes
 
    \begin{equation}
  \begin{aligned}
  \label{eq51}
 & R_{ek}=\frac{\Omega_k}{\delta^2\ln{2}}\int_{0}^{\infty} \frac{1-F_{h_{kc_k}}}{1+\frac{\Omega_k x}{\delta^2}} dx.
 \end{aligned}
\end{equation}
Substituting  $\left(\frac{1}{\Gamma(m)}\gamma(m,\frac{mx}{\lambda})\right)^{(L-\upsilon)}$
into $F_{h_{kc_k}}$ in (\ref{eq51}) completes the proof.

\section{Proof of Proposition \ref{pro2}}
\noindent The proof of  Proposition \ref{pro2} is similar to that of Proposition \ref{pro1} with the only difference that the channel  $h_{kc_k}$ is chosen randomly. Thus, the PDF and CDF of  $x=|h_{kc_k}|^2$ follow  (\ref{eq3}) and (\ref{eq41}), respectively.
Substitution  the CDF  of $x$ into (\ref{eq5}) leads to 
 
    \begin{equation}
  \begin{aligned}
  \label{eq5}
 & R'_{ek}=\frac{\Omega_k}{\delta^2\Gamma(m)\ln{2}}\int_{0}^{\infty} \frac{\Gamma(m,\frac{mx}{\lambda})}{1+\frac{\Omega_k x}{\delta^2}} dx,
 \end{aligned}
\end{equation}
and since  $\frac{1}{1+\frac{\Omega_k x}{\delta^2}} = G_{1,1}^{1,1} \left( \frac{\Omega_k x}{\delta^2}\bigg| \begin{matrix}0\\ 0 \end{matrix} \right)$ \cite[Eq. (8.4.2.5)]{newton} and  $\Gamma(m,\frac{mx}{\lambda})=G_{2,0}^{1,2} \left( \frac{m}{\lambda}x\bigg| \begin{matrix}1\\ 0, m \end{matrix} \right)$  \cite[Eq. (8.4.16.2)]{newton}, (\ref{eq5}) can be rewritten as 
\begin{equation}
\begin{aligned}
\label{erg31}
R'_{ek}= \frac{\Omega_k \int_{0}^{\infty} G_{1,1}^{1,1} \left( \frac{\Omega_k x}{\delta^2}\bigg| \begin{matrix}0\\ 0 \end{matrix} \right)G_{1,2}^{2,0} \left( \frac{m}{\lambda}x\bigg| \begin{matrix}1\\ 0, m \end{matrix} \right)dx}{\delta^2\Gamma(m) \ln{(2)}}.
\end{aligned}
\end{equation}
By using the relationship in \cite[Eq. (7.811)]{newton1},  
  (\ref{erg3}) can be obtained, which completes the proof of Proposition 2. 
\bibliographystyle{IEEEtran}
 
\bibliography{references}
\end{document}